\documentclass[runningheads]{llncs}
\usepackage{verbatim} % [GZ] for "comments"
\usepackage{graphicx}
\usepackage[pagebackref=false,breaklinks=true,letterpaper=true,colorlinks,bookmarks=false]{hyperref}

\usepackage{amssymb}
\usepackage{amsmath}
\usepackage{cite}
\usepackage{xcolor,colortbl}
\usepackage[labelfont=bf,labelsep=period]{caption}

\newcommand{\correct}{\cellcolor{green}}
\newcommand{\wrong}{\cellcolor{red}}
\newcommand{\non}{\cellcolor{gray}}

\begin{document}
\title{Decision Support for Intoxication Prediction Using Graph Convolutional Networks}

\titlerunning{ Decision Support for Intoxication Prediction}

\begin{comment} % ======== [ANONYMOUS]
\author{
$\ast\ast\ast$ \\
%$\ast\ast\ast$
}
%
\authorrunning{$\ast\ast\ast$} % [OPTION] NAMES
\institute{
$\ast\ast\ast$ \\
\email{$\ast\ast\ast$@$\ast\ast\ast$}
%$\ast\ast\ast$ \\
%$\ast\ast\ast$ \\
%$\ast\ast\ast$ \\
%$\ast\ast\ast$ \\
%$\ast\ast\ast$
}
\end{comment} % ======== [ANONYMOUS]
%
%\begin{comment} % ======== [NOT ANONYMOUS]
\author{
Hendrik Burwinkel\inst{1,}\thanks{Both authors share first authorship.}
\and
Matthias Keicher\inst{1,\star}
\and
David Bani-Harouni\inst{1}
\and
Tobias Zellner\inst{4}
\and
Florian Eyer\inst{4}
\and
Nassir Navab\inst{1,3}
\and
Seyed-Ahmad Ahmadi\inst{2}
}
% index{Burwinkel, Hendrik}
% index{Keicher, Matthias}
% index{Bani-Harouni, David}
% index{Zellner, Tobias}
% index{Eyer, Florian}
% index{Navab, Nassir}
% index{Ahmadi, Seyed-Ahmad}
%
\authorrunning{H. Burwinkel et al.} % [OPTION] NAMES

\institute{
Computer Aided Medical Procedures, Technische Universit{\"a}t M{\"u}nchen, Boltzmannstra{\ss}e 3, 85748 Garching bei M{\"u}nchen, Germany\\
\email{hendrik.burwinkel@tum.de}
\and
German Center for Vertigo and Balance Disorders, Ludwig-Maximilians Universit\"at M\"unchen, Marchioninistr. 15, 81377 M\"unchen, Germany
\and
Computer Aided Medical Procedures, Johns Hopkins University, 3400 North Charles Street, Baltimore, MD 21218, USA
\and
Division of Clinical Toxicology and Poison Control Centre Munich, Department of Internal Medicine II, TUM School of Medicine, Technische Universit\"at M\"unchen, Ismaninger Str 22, 81675 M\"unchen, Germany
}
%\end{comment} % ======== [NOT ANONYMOUS]

\maketitle              % typeset the header of the contribution
\begin{abstract}
Every day, poison control centers (PCC) are called for immediate classification and treatment recommendations if an acute intoxication is suspected. Due to the time-sensitive nature of these cases, doctors are required to propose a correct diagnosis and intervention within a minimal time frame. Usually the toxin is known and recommendations can be made accordingly. However, in challenging cases only symptoms are mentioned and doctors have to rely on their clinical experience. Medical experts and our analyses of a regional dataset of intoxication records provide evidence that this is challenging, since occurring symptoms may not always match the textbook description due to regional distinctions, inter-rater variance, and institutional workflow. Computer-aided diagnosis (CADx) can provide decision support, but approaches so far do not consider additional information of the reported cases like age or gender, despite their potential value towards a correct diagnosis. In this work, we propose a new machine learning based CADx method which fuses symptoms and meta information of the patients using graph convolutional networks. 
%These networks use structural information to build patient cohorts, and map symptoms to diagnostic labels through graph signal processing. 
We further propose a novel symptom matching method that allows the effective incorporation of prior knowledge into the learning process and evidently stabilizes the poison prediction. We validate our method against 10 medical doctors with different experience diagnosing intoxication cases for 10 different toxins from the PCC in Munich and show our method's superiority in performance for poison prediction.

\end{abstract}
\section{Introduction}
Intoxication is undoubtedly one of the most significant factors of global suffering and death. In 2016, the abuse of alcohol alone resulted in 2.8 million deaths globally, and was accountable for 99.2 million DALYs (disability-adjusted life-years) -- 4.2\% of all DALYs. Other drugs also summed up to 31.8 million DALYs and 451,800 deaths world-wide \cite{Degenhardt2018}. In case of an intoxication, fast diagnosis and treatment are essential in order to prevent permanent organ damage or even death \cite{Kulling1986}. Since not all medical practitioners are experts in the field of toxicology, specialized poison control centers (PCC) like the center in Munich were established. These institutions can be called by anyone -- doctor or layman -- to help in the classification and treatment of patients. Most of the time, the substance responsible for the intoxication is known. However, when this is not the case, the medical doctor (MD) working at the PCC has to reach a diagnosis solely based on the reported symptoms, without ever seeing the patient face to face and give treatment recommendations accordingly. Especially for inexperienced MDs, this is a challenging task for several reasons. First, the symptom description may not match the symptoms described in the literature that is used to diagnose the patients. This is exacerbated by inter-individual, regional, and inter-institutional differences in the description of symptoms when reaching the doctor. Secondly, not all patients react to intoxication with the same symptoms and they may have further confounding symptoms not caused by the intoxication but due to other medical conditions. Thirdly, meta information like age, gender, weight, or geographic location, are not assessed in a structured way.\\
Current computer-aided diagnosis (CADx) systems in toxicology do not solve these problems. Most are rule-based expert systems \cite{Darmoni1994, Batista-Navarro2010, Long2017} which are very sensitive towards input variations. Furthermore, they do not consider meta information or population context, despite their potential value in diagnosis. We propose a model that can solve both mentioned problems. By employing Graph Convolutional Networks (GCN) \cite{Defferrard2016, Kipf2016}, we incorporate the meta information and population context into the diagnosis process in a natural way using graph structures. Here, each patient corresponds to a node, and patients are connected according to the similarity of their meta-information \cite{Parisot2017}. Connecting patients in this way leads to neighborhoods of similar patients. GCNs perform local filtering of graph-structured data analogous to Convolutional Neural Networks (CNN) on regular grids. This relatively novel concept \cite{Bronstein2016} already led to advancements in medicine, ranging from human action prediction \cite{Yan2018} to drug discovery \cite{Stokes2020}. It has also been used successfully in personalized disease prediction \cite{Parisot2017, Burwinkel2019, Kazi2019}. Notably, attention mechanisms \cite{Cheng2016, Lin2017} improve filtering by weighting similarity scores between nodes based on node features, which help to compensate for locally inaccurate graph structure. We base the proposed model on Graph Attention Networks (GAT) \cite{Velickovic2018}, one of the leading representatives of this GCN-class.\\
\textbf{Contributions.} Our approach for toxin prediction leverages structured incorporation of patient meta information to significantly boost performance. We further address the issue of mismatching symptom descriptions by augmenting the GCN with a parallel network layer which learns a conceptual mapping of patient symptoms to textbook symptoms described in literature. This network branch is designed to explicitly incorporate domain prior knowledge from medical literature, and produces an alternative prediction. This stabilizes the output of the model and ensures a reasonable prediction. In a set of experiments on real PCC data, we show that our model outperforms several standard approaches. Ultimately, we compare our model to patient diagnoses made by 10 MDs on a separate real-life test set. The favorable performance of our model demonstrates its high potential for decision support in toxicology.

\section{Methodology}
\begin{figure}[t]
\includegraphics[width=\textwidth]{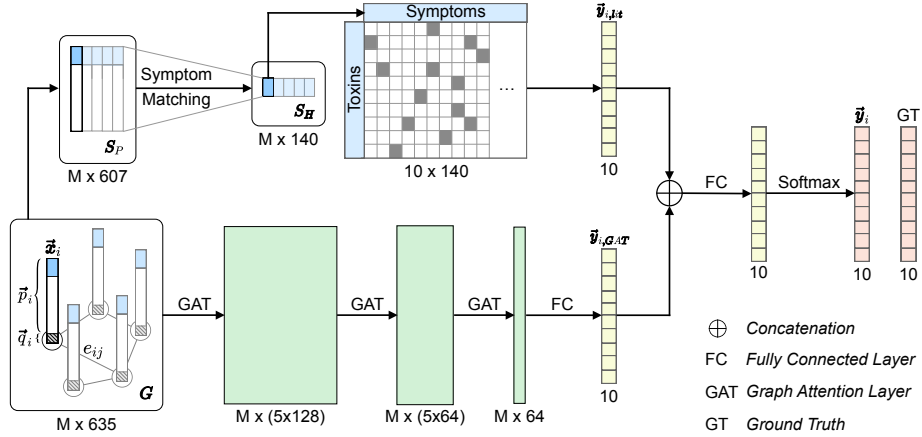}
\caption{Schematic architecture of ToxNet. The symptom vectors are processed in the graph-based GAT layers and the literature matching network in parallel.} \label{fig:flow}
\end{figure}
\textbf{General framework.} The proposed network performs the classification of the intoxication of patients with 1D symptom vectors $\textbf{P}$ using non-symptom meta information \textbf{Q} and literature symptom vectors $\textbf{H}$ in an inductive graph approach. Therefore, it optimizes the objective function $f(\textbf{P}, G(\textbf{P},\textbf{Q},\textbf{E}), \textbf{H}): \textbf{P} \rightarrow \textbf{Y}$, where $G(\textbf{P},\textbf{Q},\textbf{E})$ is a graph with vertices containing symptoms $\textbf{P}$ and meta data \textbf{Q}. Binary edges \textbf{E} denote connections between the vertices and \textbf{Y} is a set of poison classes. The symptom vectors contain a binary entry for every considered symptom, 1 if the symptom is present, 0 if not. Therefore, every patient has an individual symptom vector $\vec{p}_{i}$ with the occurring symptoms, and every poison has a vector $\vec{h}_{i}$ of literature symptoms that should occur for this poison, leading to the symptom sets: $\textbf{P} = \{\vec{p}_{1}, \vec{p}_{2},...,\vec{p}_{M}\}, \vec{p}_{i} \in \{0,1\}^{F_P}$, $\textbf{H} = \{\vec{h}_{1}, \vec{h}_{2},...,\vec{h}_{C}\}, \vec{h}_{i} \in \{0,1\}^{F_H}$, where $M$ is the number of patients, $C$ is the number of poison classes, $F_P$ and $F_H$ are the dimensions of the patient and literature symptom vector, respectively. Within $\textbf{Q} = \{\vec{q}_{1}, \vec{q}_{2},...,\vec{q}_{M}\}$, every vector $\vec{q}_{i}$ contains the patient's meta information. For every vertex in the graph we concatenate the patient symptom vector $\vec{p}_{i}$ with the meta data $\vec{q}_{i}$ of \textbf{Q} and receive \textbf{X} with vectors $\vec{x}_i$ of dimension $F$. Additionally, the edges \textbf{E} are created based on the similarity of the meta information between two patients. The network processes the patient symptoms within three GAT layers and a learned explicit literature matching layer in parallel. The resulting representations are fused to predict the corresponding intoxication.\\
\textbf{Symptom vectors.} As described above, every symptom vector corresponds to a binary encoding of all symptoms present. The dimensions $F_P$ and $F_H$ of the vectors refer to the total number of individual symptoms $S_P$ and $S_H$ that are listed within all patient cases and poison descriptions respectively. Since real patient cases also show some symptoms that are not part of the literature, $F_H < F_P$ and $S_H \subseteq S_P$. The first $F_H$ entries of every $\vec{p}_i$ correspond to $S_H$.\\
\textbf{Neighborhood generation.} The edges $E$ represent the neighborhood of every concatenated vector or vertex $\vec{x}_i$ and define which vertices $\vec{x}_j \in N_i$ should be aggregated to update the current representation of $\vec{x}_i$ within a GAT layer. The neighborhood $N_i$ of $\vec{x}_i$ is defined as the set of all $\vec{x}_j$ with $e_{ij} \in \textbf{E}$. An edge $e_{ij}$ is established when the meta information of $\vec{x}_i$ and $\vec{x}_j$ is consistent.\\
\textbf{GAT layer.} To update the representation of the vectors $\vec{x}_i$ of \textbf{X}, the GAT layer applies a shared learnable linear transformation $\textbf{W} \in \mathbb{R}^{F' \times F}$ to all $\vec{x}_i$, resulting in a new representation with dimension $F'$. For every neighbor $\vec{x}_j \in N_i$, an attention coefficient $\alpha$ is calculated using the shared attention mechanism $a$. The coefficient represents the importance of $\vec{x}_j$ for the update of $\vec{x}_i$ and is calculated as $a(\textbf{W} \vec{x}_i, \textbf{W} \vec{x}_j) = \vec{a}^T[\textbf{W} \vec{x}_i|| \textbf{W} \vec{x}_j]$, where $[~||~]$ represents the concatenation of $\textbf{W} \vec{x}_i$ and $\textbf{W} \vec{x}_j$, and $\vec{a} \in \mathbb{R}^{2F'}$ denotes a single feed-forward layer. To normalize every attention coefficient $\alpha$ and allow easy comparability between coefficients, after applying the leakyReLU activation $\sigma$, for every $\vec{x}_i$ the softmax function is applied to all coefficients corresponding to $N_i$. 
\begin{equation}
\alpha_{ij} = \frac{\exp (\sigma(\vec{a}^T([\textbf{W} \vec{x}_i|| \textbf{W} \vec{x}_j])))}{\sum_{r \in N_i} \exp (\sigma(\vec{a}^T [\textbf{W} \vec{x}_i|| \textbf{W} \vec{x}_r]))}
\end{equation}
To update $\vec{x}_i$, every feature representation $\textbf{W} \vec{x}_j$ is weighted with the corresponding $\alpha_{i,j}$ and summed up to receive the new representation $\vec{x'}_i$. The GAT network repeats this step multiple times with individually learned $\textbf{W}^k$, so-called heads, to statistically stabilize the prediction and receive individual attention coefficients $\alpha^k$. The different representations $\vec{x'}_i$ are concatenated (represented as $||$) to yield the final new representation:
\begin{equation}
    \vec{x'}_i = \Vert_{k=1}^K \sigma \left( \sum_{j \in N_i} \alpha_{ij}^k \textbf{W}^k \vec{x}_j \right)
\end{equation}
Here, $K$ is the number of used heads and $\alpha_{ij}^k$ is the attention coefficient of head $k$ for the vertices $\vec{x}_i$ and $\vec{x}_j$ \cite{Velickovic2018}.\\
\textbf{Literature symptom matching.} For every toxin class $c_i$ of all toxins $C$, the literature provides a list of commonly occurring symptoms. These are encoded in the binary symptom vector $\vec{h}_i$ for every poison. We design a specific symptom matching layer $\textbf{W}_\text{symp} \in \mathbb{R}^{F_H \times F_P}$ which learns a mapping of the patient symptom vectors \textbf{P} to the literature symptoms. This concept results in an interpretable transfer function which gives deeper insight into symptom correlations and explicitly incorporates the domain prior knowledge from literature. Due to the described setup of the symptom vectors, the first $F_H$ entries of every $\vec{p}_i$ correspond to the literature symptoms. Since these should be preserved after the matching procedure, we initialize the first $F_H$ learnable parameters of $\textbf{W}_\text{symp}$ with the unity matrix $I_{F_H}$ and freeze the diagonal during training. Like this, every symptom $s$ of $S_H$ is mapped to itself. The remaining symptoms only occurring for the patient cases are transformed into a representation of a dimension in agreement with the symptoms of the literature. As a second transformation, we create a literature layer $\textbf{W}_\text{lit} \in \mathbb{R}^{C \times F_H}$ whose $i$th row is initialized with $\vec{h}_i$ for all classes $C$ and that is kept constant during training. The resulting transformation $\vec{y}_{i, lit} = \textbf{W}_\text{lit} \cdot \sigma(\textbf{W}_\text{symp}~ \vec{p}_i)$ therefore maps the patient symptoms onto the poison classes with the explicit usage of literature information.\\
\textbf{Representation fusion.} The output of the last GAT layer is processed by a FC layer to result in $\vec{y}_{i,GAT}$ with dimension $C$. The GAT and literature representations $\vec{y}_{i,GAT}$ and $\vec{y}_{i, lit}$ are concatenated, activated and transferred through a last learnable linear transformation and a softmax function onto the class output $\vec{y}_i$.

\section{Experiments and Discussion}
\subsection{Experimental setup}
\textbf{Dataset.} The dataset consists of 8995 patients and was extracted from the PCC database from the years 2001-2019. All cases were mono-intoxications, meaning only one toxin was present and the toxin was known. We chose the following toxins: ACE inhibitors (n=119), acetaminophen (n=1376), antidepressants (selective serotonin re-uptake-inhibitors, n=1073), benzodiazepines (n=577), beta blockers (n=288), calcium channel antagonists (n=75), cocaine (n=256), ethanol (n=2890), NSAIDs (excluding acetaminophen, n=1462)) and opiates (n=879). The ten toxin groups were chosen either because they are part of the most frequently occurring intoxications and lead to a different treatment and intervention or because they have clinically distinct features, lead to severe intoxications, have a specific treatment, and should not be missed. Accordingly, the different classes are unbalanced in their occurrence since e.g. intoxication due to alcohol is a very common phenomenon. Together with the patient symptoms, additional meta information for every case is given. From the full set of available information, we use the parameters age group (child, adult, elder), gender, aetiology, point of entry and week day and year of intoxication to set up the graph structure, since these resulted in best performance.\\
\textbf{Graph setup.} Our graph is based on the described meta information for every patient. An edge $e_{ij}$ between patient $\vec{x}_i$ and $\vec{x}_j$ is established when the meta data is consistent for the medically relevant selection of parameters, i.e., the above-mentioned meta parameters. This results in a sparse graph that at the same time has more meaningful edges (the graph increases the likelihood of patients with same poisonings to become connected).\\
\textbf{Network setup.} Hyperparameters: optimizer: Adam, learning rate: 0.001, weight decay: 5e-4, loss function: cross entropy, dropout: 0.0, activation: ELU, heads: 5.\\
\textbf{Model evaluation.} First, we evaluate our network against different benchmark approaches. Then we compare the different network components within an ablation study. By disabling different parts of the network, each individual contribution is evaluated. Here, 'GAT' refers to a setup where the GAT pipeline of ToxNet is used alone, 'LitMatch' to a setup where the parallel literature-matching branch of ToxNet is used alone. Additionally, we test a sequential setting, where the literature matching is performed prior, and the learned features are transferred to the GAT (ToxNet(S)). All experiments use a 10-fold cross-validation. After proving the superiority of our method, we compare our network against the performance of 10 MDs, who are classifying the same unseen subset of the full test data as our method. This subset is divided into 25 individual cases for every MD, and 25 additional cases identical for all MDs, i.e., $250+25=275$ cases. In this setup, we are able to perform a statistical performance analysis on a larger set of cases, but also evaluate the inter-variability of the medical experts to distinguish between easy and difficult cases.

\subsection{Experimental results}
\begin{figure}[t]
    \centering
    \begin{minipage}{0.5\textwidth}
        \centering
        \includegraphics[trim={0 1.5cm 0 2cm},clip,width=\linewidth]{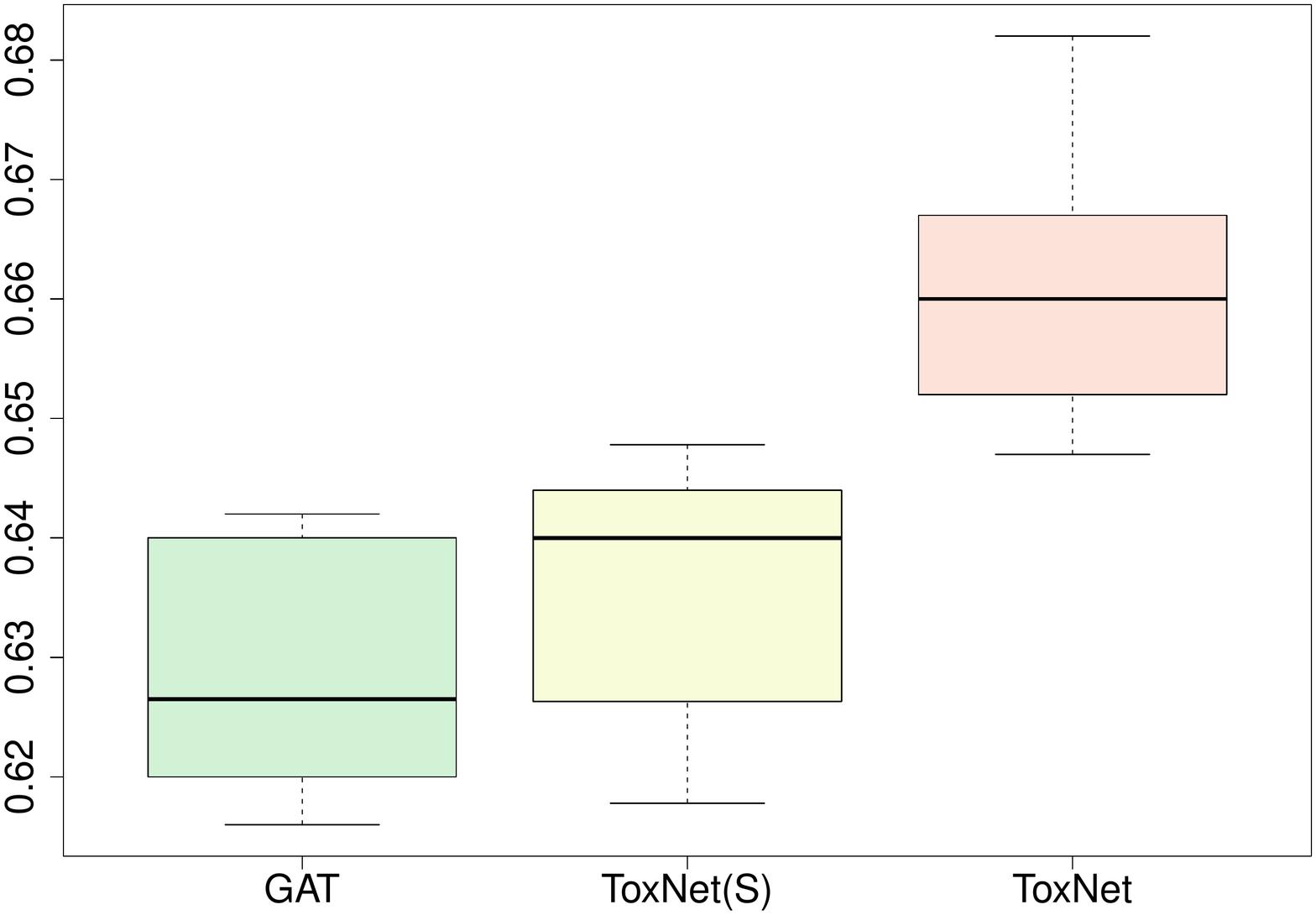}
    \end{minipage}%
    \begin{minipage}{0.5\textwidth}
        \centering
        \includegraphics[trim={0 1.5cm 0 2cm},clip,width=\linewidth]{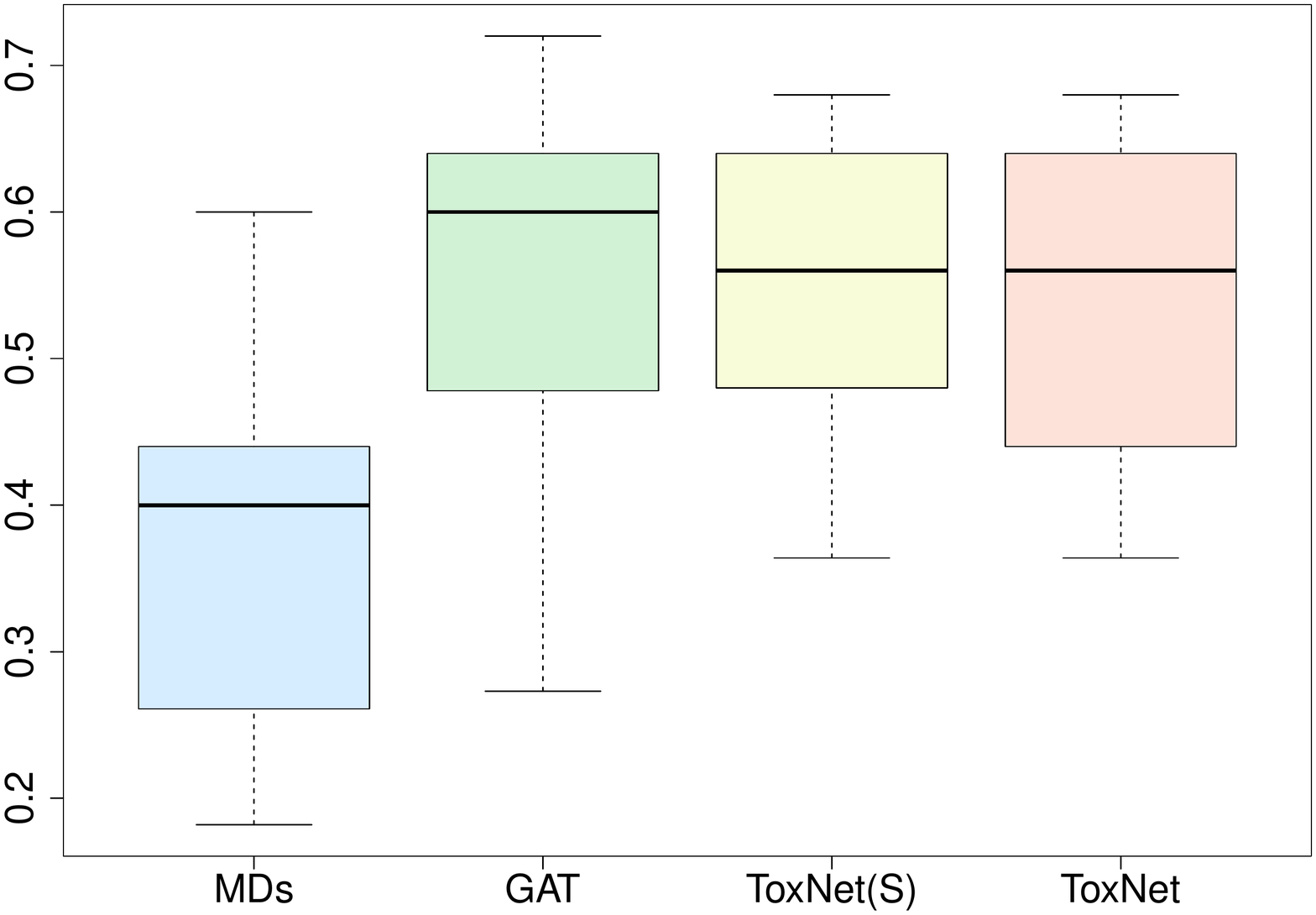}
    \end{minipage}
    \caption{\textbf{Left:} Comparison of ToxNet and different benchmark methods over 10-fold cross validation. \textbf{Right:} Comparison of ToxNet and benchmark methods with MDs' performance over 10 different sets evaluated by one MD each.}
    \label{fig:boxplots}
\end{figure}

\begin{figure}[t]
    \centering
    \includegraphics[trim = 0px 0px 0px 0px, width=1.0\textwidth]{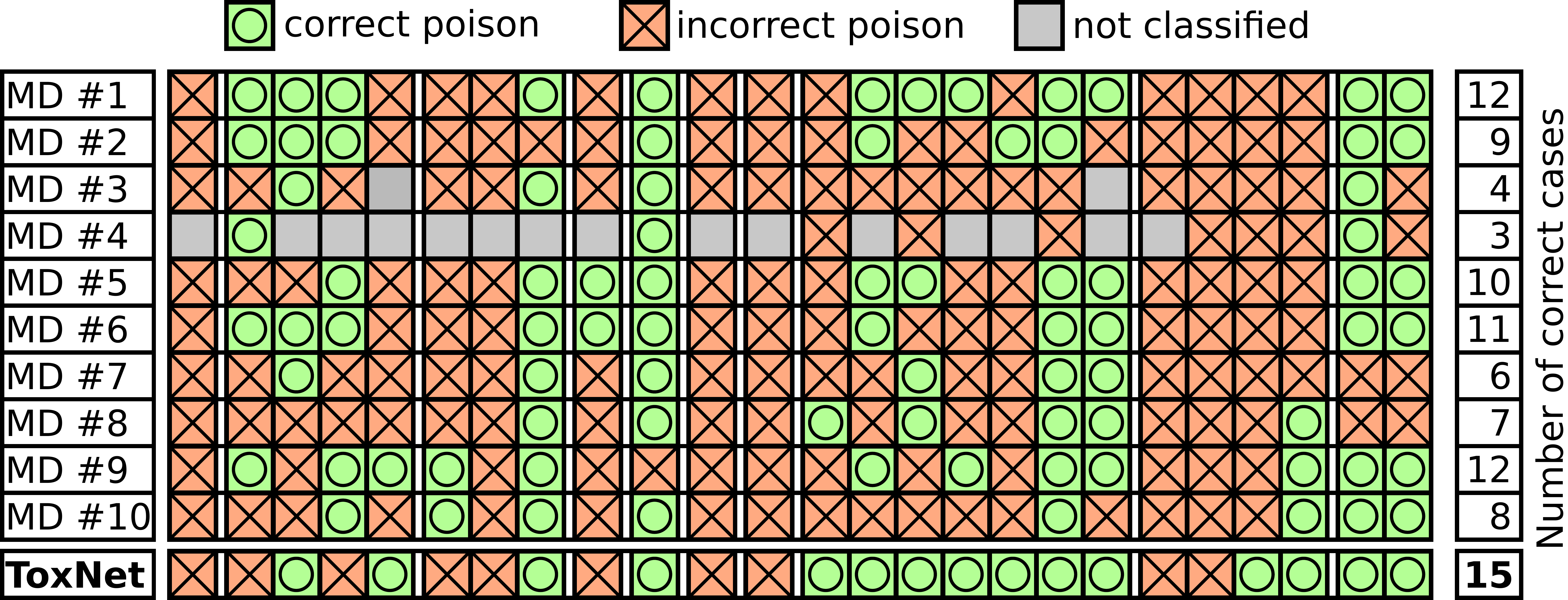}
    \caption{Clinician inter-variability and comparison with ToxNet. Poison classes are ordered alphabetically, each group separated with a white spacing.}
    \label{fig:MD_compare}
\end{figure}

\begin{table}[t]
\begin{center}
\caption{Performance comparison of different methods for poison prediction. Methods are described in detail in Sec. 3 (p-value: $<$0.01 $\ast$, $<$0.005 $\ast \:  \ast$).}
    \begin{tabular}{|p{2.8cm}|p{2.5cm}|p{2.5cm}|p{1.7cm}|p{1.7cm}|}
        \hline
        Method & F1 Sc. micro & F1 Sc. macro & p-val micro & p-val macro \\
        \hline
        Naive Matching & 0.201 $\pm$ 0.012 & 0.127 $\pm$ 0.007 & $\ast \:  \ast$ & $\ast \:  \ast$\\
        Decision Tree & 0.246 $\pm$ 0.016 & 0.227 $\pm$ 0.016 & $\ast \:  \ast$ & $\ast \:  \ast$\\
        LitMatch & 0.474 $\pm$ 0.005 & 0.342 $\pm$ 0.023 & $\ast \:  \ast$ & $\ast \:  \ast$ \\
        MLP with meta & 0.544 $\pm$ 0.015 & 0.429 $\pm$ 0.019 & $\ast \:  \ast$ & $\ast \:  \ast$\\
        %SingleLayerMatch & 0.493 $\pm$ 0.007 & 0.358 $\pm$ 0.019 & $\ast \:  \ast$ & $\ast \:  \ast$ \\
        %DoubleLayerMatch & 0.481 $\pm$ 0.009 & 0.370 $\pm$ 0.022 & $\ast \:  \ast$ & $\ast \:  \ast$ \\
        %RandomMatch & 0.474 $\pm$ 0.011 & 0.349 $\pm$ 0.022 &   $\ast \:  \ast$& $\ast \:  \ast$ \\
        %ArtificialLitMatch & 0.498 $\pm$ 0.015 & 0.345 $\pm$ 0.027 & $\ast \:  \ast$ &$\ast \:  \ast$ \\
        GAT & 0.629 $\pm$ 0.010 & 0.458 $\pm$ 0.021 & $\ast \:  \ast$ & $\ast \:  \ast$\\
        \hline
        ToxNet(S) & 0.637 $\pm$ 0.013 & 0.478 $\pm$ 0.023 & $\ast \:  \ast$ & $\ast \:  \ast$ \\
        \textbf{ToxNet} & \textbf{0.661} $\pm$ 0.010 & \textbf{0.529} $\pm$ 0.036 & / & / \\
        \hline
    \end{tabular}
\label{tab:compare}
\end{center}
\end{table}
\textbf{Performance comparison against other methods.} In Tab. \ref{tab:compare}, we compare the F1 micro and macro scores of different benchmark approaches against our method ToxNet. The Naive Matching provides a lower baseline by simply voting for the poison which has the most overlap between literature and patient symptoms. The decision tree was trained based on the literature symptoms and then used on the patient symptoms. Both models perform poorly, which leads to the conclusion that the available literature alone is not a good guide for poison classification. With the LitMatch neural network branch from our approach, we maintain the possibility to incorporate literature knowledge explicitly, but receive significantly better results. In the next step, a Multi-Layer Perceptron (MLP) with 3 hidden layers and $5 \cdot 128$, $5\cdot 64$ and $64$ hidden units respectively (same as GAT) was trained on the patient data to perform the prediction. In order to allow for a fair comparison, the patient's meta information was concatenated to their symptom vector, thus resulting in both the MLP and GAT using the same information. By comparing the MLP to a standard GAT network, it is observable that the usage of the meta information inside our graph method significantly boosts the classification performance, showing the value that the graph structure adds to the evaluation. Adding the literature information into the method by applying our proposed method ToxNet increases this performance even further. It needs to be stated that this enhancement is reached, although the literature data alone was shown not to be very informative for the task at hand. We therefore assume that there is a synergy effect, and an improvement of the literature might lead to an even stronger boost. To identify the individual contributions, both pipelines within ToxNet (GAT and LitMatch) are also evaluated separately as described above. Within our experiment, we found that the parallel setting of ToxNet is slightly superior to a sequential setting (ToxNet(S)). The results described above are also illustrated in the boxplot in Fig. \ref{fig:boxplots} (left).
\\
\textbf{Performance comparison against medical experts.} In order to evaluate the performance of our method against medical experts, we conducted a survey with 10 medical doctors (MDs) from the toxicology department of the Klinikum rechts der Isar in Munich, where each MD had to classify 50 intoxication cases that were split up as described above. Fig. \ref{fig:boxplots} (right) shows a box plot of the performance of the 10 MDs compared with different benchmark methods as well as our method ToxNet on the ten individual sets of 25 cases each, so 250 in total. All three graph-based approaches clearly outperform the MDs due to the optimized usage of meta information. For this small subset of the full test set, the performance boost of ToxNet compared to GAT is not as severe as for the full test set. However, the overall performance is more stable (smaller margins). In Fig. \ref{fig:MD_compare}, we performed a detailed inter-variability study on the 25 cases evaluated by all doctors. Except for one case, every intoxication case correctly classified by the majority of MDs, our method accomplished as well. Furthermore, for eight cases, where only half of the MDs or less correctly predicted the intoxication, our method still succeeded. These results demonstrate that our proposed ToxNet architecture can predict simple cases reliably and at an expert-level performance, while additionally providing a high prediction stability on cases that are challenging to a majority of doctors. Even compared to the two best MDs, who correctly classified 12 cases, our method overall resulted in 15 correct poison predictions. Six cases were wrongly classified by all doctors and our method. These are data samples with insufficient documentation quality (e.g. only a single generic symptom) which indicate intrinsic challenges from medical data in the wild.
%Due to varying documentation quality regarding the patient-specific symptoms, the data contained multiple challenging cases, which can explain this behavior.
%------------------------
%-------------------------

\section{Conclusion}
In this work, we proposed ToxNet, a new architecture for improved intoxication prediction. The network effectively incorporates patient symptoms, meta-information like age group or residence, and domain prior knowledge from literature. We showed that the usage of meta-data within the graph structure of a graph convolutional network inside ToxNet leads to a significantly higher classification performance than all other methods investigated. In our benchmark study, we explicitly showed that a simple concatenation of the meta-data to the patient symptom vector is not sufficient -- the improvement can be attributed to the patient graph. Additionally, we introduced a symptom matching method that allows the explicit usage of literature knowledge and included it into a parallel learning approach which further improved the overall network performance. Although we found that the literature information by itself was not informative enough for a satisfactory classification, we showed that a parallel integration with our graph network still led to synergy effects and an improved classification. We evaluated our network against 10 MDs with different experience levels and found a more stable prediction on both simple and highly challenging intoxication cases, given the high inter-rater variability among experts. We thus demonstrated the potential of ToxNet as a clinical decision support in this highly critical domain of medical intervention. On a wider scale, we believe that our architecture and validation provide a valuable case study: medical expertise can be regionally flavored and affect symptoms in a way that is not necessarily covered by expert literature. A proper modeling of these effects, fused with recent advances in graph-based population models, can lead to significant improvements in the field of computer-aided diagnosis and support clinical practice.\\
\\
%In future work, updating the literature might improve the performance of our network ToxNet even further.
\textbf{Acknowledgements}\\
%$\ast\ast\ast$
The study was supported by the Carl Zeiss Meditec AG in Oberkochen, Germany, and the German Federal Ministry of Education and Research (BMBF) in connection with the foundation of the German Center for Vertigo and Balance Disorders (DSGZ) (grant number 01 EO 0901).

%\bibliographystyle{splncs04}
%\bibliography{References/bibliography}
\bibliography{AI_Toxicology_Burwinkel}
\end{document}